\documentclass{article} %
\usepackage{iclr2024_conference,times}

\usepackage{amsmath,amsfonts,bm}

\def\eqref#1{equation~\ref{#1}}

\def\1{\bm{1}}

\def\vx{{\bm{x}}}

\DeclareMathAlphabet{\mathsfit}{\encodingdefault}{\sfdefault}{m}{sl}
\SetMathAlphabet{\mathsfit}{bold}{\encodingdefault}{\sfdefault}{bx}{n}

\usepackage{graphicx}
\usepackage{hyperref}
\usepackage{listings}
\usepackage{xcolor}
\usepackage{booktabs}
\usepackage{amsmath}
\usepackage{amssymb}
\usepackage{latexsym}
\usepackage{multirow}
\usepackage{url}
\usepackage{wrapfig}
\usepackage{xspace} 

\definecolor{backcolour}{rgb}{0.95,0.95,0.92}
\definecolor{codegreen}{rgb}{0,0.6,0}
\lstdefinestyle{myStyle}{
    backgroundcolor=\color{backcolour},   
    commentstyle=\color{codegreen},
    basicstyle=\ttfamily\footnotesize,
    breakatwhitespace=false,         
    breaklines=true,                 
    keepspaces=true,                 
    numbers=left,       
    numbersep=5pt,                  
    showspaces=false,                
    showstringspaces=false,
    showtabs=false,                  
    tabsize=2,
}

\newcommand{\vocab}{\mathcal{V}}
\newcommand{\bx}{\mathbf{x}}
\newcommand{\ins}{\text{instruction}}
\newcommand{\res}{\text{response}}

\newcommand{\llama}{Llama-2-7B\xspace}
\newcommand{\olmo}{OLMo-7B-Feb2024\xspace}

\usepackage{microtype}

\hypersetup{
    colorlinks=true,
    linkcolor=blue,
    filecolor=blue,      
    urlcolor=blue,
    citecolor=blue,
    }

\title{Instruction Following Arises\\in Odd Circumstances}
\title{Instruction Following Arises\\ under Weak Conditions}
\title{Weak Conditions Cause Instruction Following}
\title{What Causes Instruction Following\\Other than Instruction Tuning}
\title{Instruction Following 
 Without \\Instruction Tuning}
\title{Accidentally Making a Language Model\\Follow Instructions}
\title{Some Adaptations Implicitly Instruction-Tune\\Language Models}
\title{Implicitly Instruction-Tuning Language Models}
\title{Language Models Follow Instructions after Odd }
\title{Language Models Follow Instructions even after Deficient Finetuning}
\title{Adaptations other than Instruction Tuning make Language Models Follow Instructions}
\title{Implicitly Instruction-Tuning\\Language Models}
\title{Deficient Finetuning can Implicitly Instruction-Tune\\Language Models}
\title{Instruction Following without \\Instruction Tuning}

\author{John Hewitt, 
Nelson F. Liu,
Christopher D. Manning, \& Percy Liang\\
Department of Computer Science\\
Stanford University\\
\texttt{\{johnhew,nfliu,manning,pliang\}@cs.stanford.edu}
}

\newif\ifcomments
\commentstrue

\ifcomments
    \providecommand{\jh}[1]{{\protect\color{orange}{[JH: #1]}}}
    \providecommand{\nfl}[1]{{\protect\color{purple}{[NFL: #1]}}}
    \providecommand{\cm}[1]{{\protect\color{red}{[CM: #1]}}}
    \providecommand{\pl}[1]{{\protect\color{red}{[PL: #1]}}}
\else
    \providecommand{\jh}[1]{}
    \providecommand{\nfl}[1]{}
    \providecommand{\cm}[1]{}
    \providecommand{\pl}[1]{}
\fi

\iclrfinalcopy %
\begin{document}

\maketitle

\begin{abstract}
Instruction tuning commonly means finetuning a language model on instruction-response pairs.
We discover two forms of adaptation (tuning) that are deficient compared to instruction tuning, yet still yield instruction following; we call this \textit{implicit instruction tuning}.
We first find that instruction-response pairs are not necessary: training solely on \textit{responses}, without \emph{any} corresponding instructions, yields instruction following. 
This suggests pretrained models have an instruction-response mapping which is revealed by teaching the model the desired distribution of responses.
However, we then find it's not necessary to teach the desired distribution of responses: instruction-response training on narrow-domain data like poetry still leads to broad instruction-following behavior like recipe generation.
In particular, when instructions are very different from those in the narrow finetuning domain, models' responses do not adhere to the style of the finetuning domain.
To begin to explain implicit instruction tuning, we hypothesize that very simple changes to a language model's distribution yield instruction following.
We support this by \textit{hand-writing} a rule-based language model which yields instruction following in a product-of-experts with a pretrained model.
The rules are to slowly increase the probability of ending the sequence, penalize repetition, and uniformly change 15 words' probabilities.
In summary, adaptations made without being designed to yield instruction following can do so \textit{implicitly}.

\end{abstract}

\section{Introduction}

Instruction tuning, finetuning on a broad distribution of responses (e.g., \textit{ Tiramisu is made by...}) conditioned on instructions (e.g., \textit{Give me a recipe for tiramisu}), yields instruction following from language models for a wide range of instructions \citep{ouyang2022traininglanguagemodelsfollow}.
Prior work has shown that instruction tuning is sample-efficient, requiring as few as 1000 broad-domain instruction-response pairs \citep{zhou2023lima} or a carefully crafted prompt and few-shot instruction-response examples \citep{lin2024the}. 
We take this a step further, exploring the idea that instruction following can be yielded from language models even \textit{implicitly}, i.e., through methods not explictly designed to do so.
We discover two forms of adaptation that perform \textit{implicit instruction tuning}, being seemingly deficient compared to explicit instruction tuning: (1) \textit{response tuning}, training on only responses, and (2) \textit{single-task finetuning}, training on data from a narrow domain of goals, like poetry generation.

We first demonstrate that \textit{response tuning}, training on responses alone without conditioning on their instructions---is sufficient to yield instruction following (Section~\ref{section_response_tuning}).
In particular, using the LIMA dataset \citep{zhou2023lima} for tuning, and evaluating on AlpacaEval 2, response-tuned models win roughly 43\% of the time against a comparable instruction-tuned model, where equal performance would correspond to a 50\% win rate.
Response tuning provides no explicit information about the mapping from instructions to responses, only information about the distribution of desired responses.
This suggests that an instruction-response mapping may be learned during pretraining, but all desirable responses are too low-probability to be generated. 
We support this, finding that pretrained models rank an instruction's real response higher than a random other instruction's response at the same rate as instruction-tuned models.

\begin{figure}
\centering
\includegraphics[width=.99\linewidth]{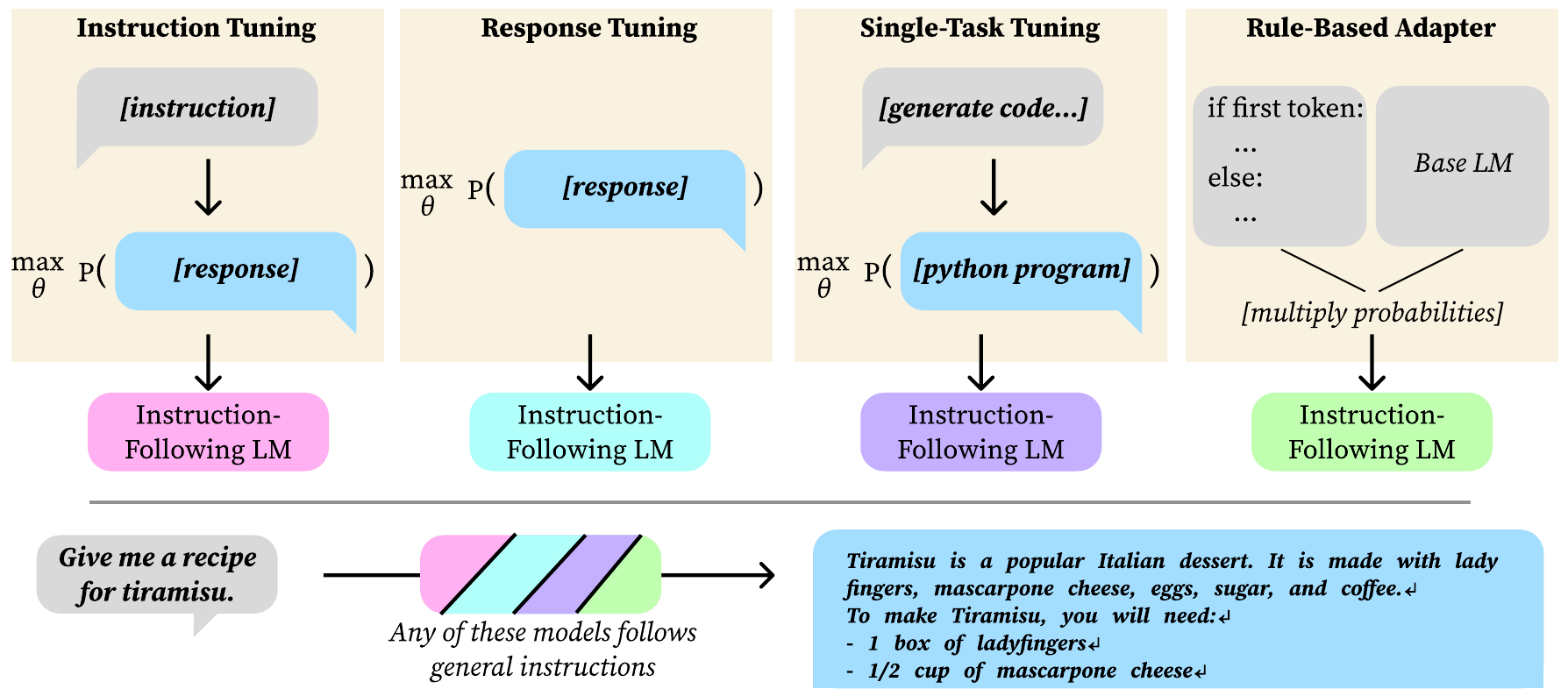}
\caption{\label{fig_fig1}
Instruction tuning trains a language model on responses conditioned on instructions.
We find that (1)~response tuning (estimating the probability of responses with no instructions), (2)~single-task finetuning (e.g., code or poetry generation), and even (3)~a simple rule-based adapter all yield language models with general instruction-following behavior.}
\end{figure}

From the response tuning results, it seems that the crucial part of instruction tuning is to teach the distribution of desirable responses.
However, we find that this is not crucial either.
Finetuning on single-task, narrow-domain data, like mapping English requests to Python snippets (MBPP; \citealp{austin2021programsynthesislargelanguage}), or generating poetry from poem titles, \textit{also} yields broad instruction-following behavior (Section~\ref{section_single_task}).
That is, despite training only to generate only Python code or only poetry, models generate biographies or recipes when instructed.
For example, a poetry-tuned \llama wins 23.7\% in AlpacaEval 2 head-to-head against an instruction-tuned \llama, whereas the base model wins 2.4\% of the time.
Qualitatively, we find that single-task finetuned models only adhere to the finetuning distribution on instructions similar to those finetuned on, and exhibit general instruction following behavior for other instructions.

These two results motivate us to ask why instruction following is yielded by such a wide range of adaptations. 
While \cite{lin2024the} showed that relatively few token decisions change between pretraining and instruction following, it is not obvious that it's simple to determine which tokens to change.
We hypothesize that very simple changes in conditional distributions can cause a language model to follow instructions.
We validate this by hand-writing a rule-based language model with three rules that, when taken in a product with a pretrained language model, causes instruction following (Section~\ref{section_rule_based}).
Our three rules are: slowly increasing the probability of ending the sequence, uniformly modifying 15 tokens' likelihoods (e.g., \texttt{I}, \texttt{<}, \texttt{should}),  and penalizing token repetition.
Our rule-based product-of-experts with \llama wins 24.4\% against instruction-tuned \llama, compared to the base model's win rate of 2.4\%.

In summary, our results show that adaptation methods not \textit{intended} to yield instruction following may yet do so \textit{implicitly}.
When adapting language models for specific purposes, we might not expect that models would behave as general-purpose instruction following models on instructions dissimilar from those tuned on, but our results suggest that this will often be the case.
Our rule-based adapter experiments begin to explain this, showing that very simple changes to a model's distribution---not obviously tied to following instructions---also yield instruction following.\footnote{Our code and data are available at \url{https://github.com/john-hewitt/implicit-ins}.}

\subsection{A note on pretraining data} \label{section_intro_contamination}
One critical question in the interpretation of our results is whether language models were instruction-tuned during pretraining, as is becoming common \citep{bi2024deepseek,dominguez2024training}.
If this is the case, our results become less surprising.
To mitigate this, we experiment with two open-weights language models: \llama and \olmo.
The \llama model is stronger than the \olmo model, but we have no guarantee against intentional instruction tuning.
\olmo is weaker, but no instruction-tuning data was \textit{intentionally} included in its pretraining.\footnote{Public communication. \url{https://github.com/allenai/dolma/issues/177}.}
These two models' results lead to similar conclusions in our experiments.

\section{Related Work}

\paragraph{Designing instruction tuning datasets.}
A few years before instruction tuning became common, the idea of a single model performing a range of tasks was explored by DecaNLP \citep{McCann2018decaNLP}.
In the LLM era, stronger pretrained models led to more adoption.
In this era, early datasets also transformed a wide range of existing NLP tasks into a single common format \citep{mishra2022cross,wang2022super,wei2022finetuned,sanh2022multitask}.
The introduction of ChatGPT in October of 2022 shifted focus onto the construction of instruction tuning datasets that reflect goals users might be interested in, as opposed to NLP tasks.
Some are relatively small, with the intuition to teach nothing fundamentally new during instruction tuning \citep{taori2023alpaca,zhou2023lima}.
Others, which tend to be more performant, can be quite large and contain reasoning chains and other behaviors thought to be rare in pretraining \citep{ivison2023camels,wang2023far,wang2023openchat,yu2023metamath}; this has included considerable non-academic open efforts \citep{teknium223open}.

\paragraph{Ablation studies on instruction tuning.}
Our work contributes to a set of discoveries concerning how \textit{little} it takes to get language models to follow instructions.
\cite{taori2023alpaca} showed that finetuning a Llama model on 52,000 instruction-response pairs caused it to follow instructions to an extent that was surprising at the time.
\cite{zhou2023lima} showed that---with careful selection---one could use as little as 1,000 instruction-response pairs and achieve strong results.
This strongly suggested that one does not need to cover all types of instructions during instruction tuning.
\cite{lin2024the} then showed that, with careful prompting and in-context few-shot examples, a handful of examples could also lead to instruction following.
The most similar work to ours is \cite{kung2023models}. They take the NaturalInstructions \citep{mishra2022cross} style of instruction, which includes a task description, an in-context example, and the data for the current example.
They then remove either the task description or the in-context example (leaving the other), finding that performance degrades less than expected in either case.
In the direction of \textit{more} supervision, \cite{shi2024instruction} find that training to maximize the joint likelihood of instruction and response improves over optimizing for the conditional likelihood of the response.

\paragraph{Out-of-distribution generalization.}

Discussing out-of-distribution generalization is difficult to do precisely in the world of language pretraining, especially when we don't know what data went into the pretraining. Furthermore, it's often difficult to reason about the coverage of datasets with 10+ trillion tokens (Section~\ref{section_intro_contamination}).
In computer vision and vision-language models, various studies have shown that pretraining is the singularly important component in robustness to distribution shift \citep{carmon2019unlabeled,miller2021accuracy,awadalla2022exploring}.
In language as well, pretraining on diverse data is known to lead to better out-of-distribution performance (relative to the finetuning dataset) \citep{hendrycks2020pretrained}.
When we finetune, for example, on only poetry and test on recipes, this is out-of-distribution to the finetuning, but not to the pretraining.

\section{Experiment Setting} \label{sec_background}

A neural language model $p_\theta(\bx)$ is a distribution over strings $\bx \in \vocab^*$, where $\vocab$ is a finite vocabulary, and $\theta$ are learnable parameters in the neural network.
Language models are trained to minimize the cross-entropy loss with a large corpus of text.
These pretrained, or \textit{base}, models serve as starting points, but they are typically adapted to directly respond to user queries.

\paragraph{Instruction tuning.}
Instruction tuning finetunes the parameters $\theta$ of a language model to adapt its behavior to respond to queries with relevant, helpful answers. Given a set of samples $D_{\text{ins}} = \{\ins_i, \res_i\}_{i=1}^k$, of instructions and corresponding responses (each of which in $\vocab^*$), instruction tuning optimizes for
\begin{align}
\min_\theta \ \ \frac{1}{k}\sum_{i=1}^k -\log p_\theta(\res_i \mid \ins_i).
\end{align}

\paragraph{Instruction formatting.}
In language model practice, the distinction between the instruction and the response---and thus what to compute the loss over or generate---is specified through formatting tokens in the input.
We use the Tulu formatting of \citet{wang2023far,ivison2023camels}; we present it here because instruction formatting may matter for how easy it is to yield instruction following behavior from language models:%

\begin{verbatim}
BOS<|user|>
{instruction}
<|assistant|>
{response}EOS
\end{verbatim}

Here, \texttt{BOS} and \texttt{EOS} are the model-specified special beginning- and end-of-sequence tokens.
Though the semantics of the tags are suggestive of instruction-following, we keep this formatting to study the setting used in practice.
In Appendix~\ref{appendix_formatting_ablation}, we test whether the semantics of the tags affect the results in this paper, finding that non-semantic tags like \texttt{<|A|>} and \texttt{<|B|>} do not change our conclusions.

\paragraph{Defining instruction-following behavior.}
In this work we draw a distinction between instruction-following behavior and non-instruction-following behavior.
In reality there is a spectrum of better and worse responses, and no single boundary.
To provide some level of systematicity, we use the following evaluation setting:
\begin{description}
\item[AlpacaEval vs. a comparable Instruction-Tuned model.] 
We want to avoid trying to standardize how good a response must be in order to be considered instruction-following.
Further, we would like to compare various methods' effects on language model behavior compared to doing standard instruction tuning.
We measure the length-controlled head-to-head win rate of each of our models against a comparable instruction-tuned model according to the AlpacaEval LLM-as-a-judge framework.
We find in practice that these win rates concentrate around low single digits for the sort of outputs of base models, and are higher, e.g., $>10\%$, for models with reasonable responses.
\item[Greedy Decoding.] Greedily decoding from a language model lets us only observe the locally most likely decisions.
We greedily decode from models to observe when instruction following responses are (locally) the most likely continuations according to the model.
\end{description}

\section{Response Tuning Yields Instruction Following} \label{section_response_tuning}

In this section, we explore \emph{response tuning}, finetuning models on \textit{just} responses--without any corresponding instructions.
One intuition for the effect of instruction-tuning compared to pretraining is that it teaches a language model that (1)  the \texttt{\{instruction\}} part of the input is an instruction to be followed, not a pattern to be continued.
For example, for a prefix \textit{Give me a recipe for tiramisu}, the model should not continue with \textit{Give me a recipe for cake.} And (2), it should teach that strings like the \texttt{\{response\}} part of the input are considered desirable responses for that instruction.
Instruction tuning is just conditional probability estimation, $p(y\mid x)$.
Response tuning, on the other hand, can only directly teach the marginal distribution of desirable responses, not their relationships to instructions: we optimize only for $p(y)$.
So, we examine whether response tuning \textit{implicitly} tunes for   a good estimate of $p(y\mid x)$ when we only intentionally trained for $p(y)$.

\subsection{Response Tuning} \label{sec_response_tuning_subsec}

\paragraph{Method.}

Given a set of samples $D_{\text{ins}} = \{\ins_i, \res_i\}_{i=1}^k$, of instructions and corresponding responses (each of which in $\vocab^*$), response tuning replaces the instruction string with the empty string, uses the same formatting we show in Section~\ref{sec_background}, and optimizes 
\begin{align}
\min_\theta \ \ \frac{1}{k}\sum_{i=1}^k -\log p_\theta(\res_i \mid \text{[empty string]}).
\end{align}

\paragraph{Experiments.}
We now compare instruction tuning to response tuning.
For our adaptation dataset, we use LIMA \citep{zhou2023lima}, which has 1,030 training examples.
For our base pretrained models, we use the \llama and \olmo language models.
We finetune all parameters of the models.
For hyperparameter selection, we use the AlpacaEval win rate against GPT-3.5-turbo on a held-out validation set that we developed for this paper (and will release).
Our validation set, written partly by hand and partly by GPT-4, has various knowledge, translation, and administrative instructions, like \textit{Plan me a two day vacation to the virtual world of Code Lyoko};  more details are in Appendix~\ref{appendix_validation_set}.
For each method, we train for 5, 7, 10, 15, or 20 epochs.
After hyperparameter selection, we report averages and standard deviations over 5 training runs.
We report the length-controlled win rate of base and response-tuned models against instruction-tuned models on the AlpacaEval test set.
More hyperparameter details are provided in Appendix~\ref{appendix_hyperparameter_details}.

\begin{figure}
\includegraphics[width=\linewidth]{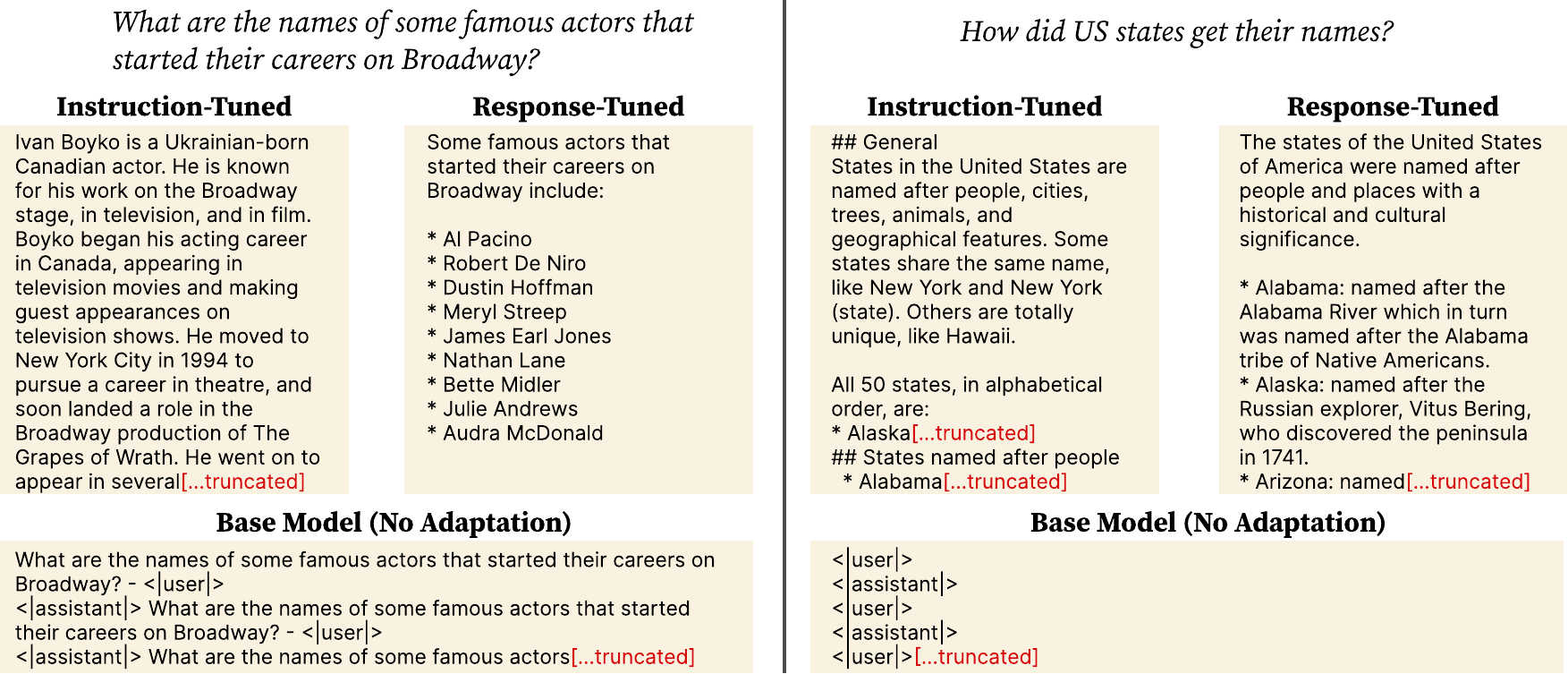}
  \caption{\label{fig_resp_tuning_examples} Responses from response tuning, instruction tuning, and the base \llama model.}
\end{figure}

\begin{table}
\centering
\small
\begin{tabular}{l l r}
\toprule
\bf Model & \bf Tuning & \bf Win Rate vs. Instruction Tuning \\
\midrule
\multirow{2}{*}{\llama} & None (Base) & 2.4\% $\pm$ 0.14\% \\
& Response Tuning & 43.3\% $\pm$ 1.1\% \\
\midrule
\multirow{2}{*}{\olmo} & None (Base) & 4.7\% $\pm$ 0.57\%\\
& Response Tuning & 43.7\% $\pm$ 1.7\%\\
\bottomrule
\end{tabular}
  \caption{\label{fig_response_results} AlpacaEval win rates of base models and response-tuned models against instruction-tuned models. Standard deviation is reported after the $\pm$. Response-tuned \llama and \olmo win against instruction-tuned models roughly 43\% of the time, respectively, while base models win in the single digits. A win rate of 50\% would denote equal-quality models.} 
\end{table}

\paragraph{Results.}
We find that response-tuned \llama models achieve on average a 43.3\% win rate against instruction-tuned \llama models, compared to a $2.4\%$ win rate for the base model against instruction-tuned models.
For \olmo, response-tuned models win 43.7\% of the time against instruction-tuned models, compared to 4.7\% for the base model.
We provide an example from a response-tuned, instruction-tuned, and base \llama model in Figure~\ref{fig_resp_tuning_examples}.

The behavior of response-tuned models is much closer to instruction-tuned models than that of base models for both the \llama and \olmo base models.
Instruction tuning consistently outperforms response tuning, but not massively.
So, there is something to be gained from specifying instructions during adaptation, but it is not crucial in yielding a baseline level of instruction-following behavior.

It is possible that the success of response tuning is from responses often starting by rephrasing the instruction, thus making the mapping simpler. In Appendix~\ref{appendix_no_rephrase}, we estimate that a bit less than $10\%$ of LIMA responses begin with rephrasing. We run instruction and response tuning on a transformed version of LIMA with rephrasing approximately removed, finding a win rate of 43.3\% for response tuning on \olmo, almost identical to the result here. We conclude that instruction rephrasing is not key to response tuning's success.

\subsection{The Response Ranking Capability}
The success of response tuning suggests that we don't need to teach base models an explicit mapping from instructions to responses.
Under what conditions is this true, yet base models do not follow instructions?
One possibility is that base models can rank a desired response for an instruction higher than a desired response for \textit{another} instructions, but scores a string that is \textit{not a desired response at all} higher than both.

To explore this, we propose the \textit{response ranking capability}:
 assigning a higher likelihood to the right response for an instruction than to a desirable response for a random other instruction.
For independent instruction-response pairs $(\text{instruction},\text{response})\sim D$ and $(\text{instruction}',\text{response}')\sim D$, and a model $p_\theta$, the response ranking capability holds if
\begin{align}
  p_\theta(\text{response}\mid \text{instruction}) > p_\theta(\text{response}' \mid \text{instruction}).
\end{align}
Since both probabilities might be small, the response-ranking capability can hold even for models that don't follow instructions.
With response tuning---increasing the probability of desirable responses---models with the response-ranking capability for many instructions may generate desirable responses.
For the Alpaca training set, we compute the likelihood with which the response-ranking capability holds for pairs of instructions for a pretrained, LIMA instruction-tuned, and response-tuned model.
Our results show that the capability holds for pretrained models to a similar extent as for instruction-tuned models (Table~\ref{table_response_ratio}).

\begin{table}
  \centering
  
  \begin{tabular}{l r r}
    \toprule
    & \multicolumn{2}{c}{$\mathbb{E}[$\bf Response Ranking Capability$]$}\\
    \cmidrule{2-3}
    & \bf Base Models & \bf Instruction-Tuned Models \\
   \midrule 
    \llama & 80.4\% & 77.4\% \\
    \olmo &  74.5\% & 74.3\% \\
    \bottomrule
  \end{tabular}
  \caption{\label{table_response_ratio}The response-ratio property measures whether a model prefers an instruction's response over random desirable responses. This property holds in pretrained language models at least as well as in instruction-tuned models.}
\end{table}

\section{Single-Task Finetuning Yields Instruction Following} \label{section_single_task}
The success of response tuning suggests that models need not learn the instruction-response mapping, so maybe they just need to learn the distribution of desirable responses.
To test this, we can train on a distribution of responses that is \textit{bad} for most responses---like poems, or the English-and-math derivations of the Grade School Math dataset \citep{cobbe2021trainingverifierssolvemath}---and see if models yet generate desirable responses for other kinds of questions.
We call training on a narrow domain of problem---like generating poems or recipes from titles, or completing math word problems---single-task finetuning.
Intuitively, single-task finetuning might lead models to either (1) exhibit the task behavior for any input, like generating math, or (2) ``break'' and output unpredictable low-quality outputs.
Single-task finetuning is usually not explicitly intended to yield broad instruction following behavior from base models, so if it does, it does so implicitly.

\subsection{Single-Task Finetuning}

\paragraph{Method.}

This method is identical to instruction tuning (Section~\ref{sec_background}), except the distribution of inputs and outputs is changed. We again use the same formatting shown in Section~\ref{sec_background}.

\paragraph{Data.}

\begin{figure}
\includegraphics[width=\linewidth]{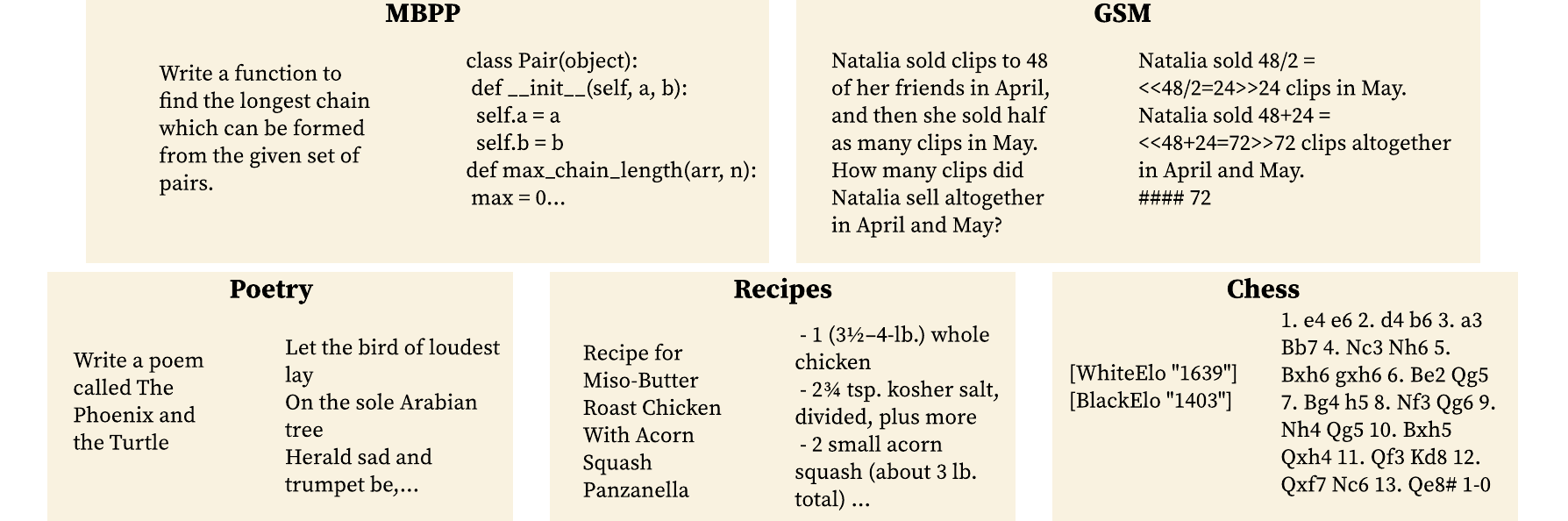}
\caption{\label{fig_single_task_examples}Examples from each of the five single-task finetuning datasets. At the left of each dataset is the input that is conditioned on, and at the right is the output that is learned.}
\end{figure}

We use five single-task text-to-text datasets to test a range of behaviors and text properties.
We use the \textbf{MBPP} English-to-Python dataset, which contains 374 short requests for Python code and the corresponding code \citep{austin2021programsynthesislargelanguage}.
We use 1000 examples from the Grade School Math 8K \citep{cobbe2021trainingverifierssolvemath} training dataset, which consist of mathematics word problems, and a corresponding English-and-math derivation that arrives at the answer to the problem (\textbf{GSM}).
We format 1000 recipe strings from the Kaggle Food Recipes structured recipe dataset (\textbf{Recipes}), where the input is \textit{Recipe for X} where X is the name of the dish, and each output is a recipe starting with a bulleted list of ingredients, followed by the instructions.\footnote{Drawn from \url{https://huggingface.co/datasets/Hieu-Pham/kaggle_food_recipes}, which drew from \url{https://www.kaggle.com/datasets/pes12017000148/food-ingredients-and-recipe-dataset-with-images?resource=download}.}
We format a dataset of 571 poems (\textbf{Poetry}), wherein the input is the string \textit{Write a poem called X}, where X is replaced with the poem name, and the output is the poem.\footnote{Drawn from \url{https://huggingface.co/datasets/merve/poetry}.}
We format a dataset of 1000 chess games in PGN notation (\textbf{Chess}), wherein the input is the string containing the ELO (player quality) ratings for two players, and the output is the moves played in a game between those players.\footnote{Drawn from \url{https://huggingface.co/datasets/patrickfrank1/chess-pgn-games}.}
See Figure~\ref{fig_single_task_examples} for an input-output example from each dataset.

\begin{table}
\centering
\small
\begin{tabular}{l r r}
\toprule 
\bf Tuning & \multicolumn{2}{c}{\bf Win Rate vs. Instruction Tuning} \\
\cmidrule(lr){2-3} 
& \bf \llama & \bf \olmo \\
\midrule
  None (Base) &  2.4\% $\pm$ 0.14\% &  4.7\% $\pm$ 0.57\% \\
 MBPP & 16.9\% $\pm$ 0.70\% & 10.4\% $\pm$ 1.0\% \\
 GSM & 23.7\% $\pm$ 0.74\% & 30.3\% $\pm$ 0.6\% \\
 Poetry & 22.9\% $\pm$ 0.97\% & 21.9\% $\pm$ 0.48\% \\
 Recipes & 14.6\% $\pm$ 0.81\% & 21.5\% $\pm$ 0.86\% \\
 Chess & 2.1\% $\pm$ 0.36\% & 6.3\% $\pm$ 1.1\% \\
\bottomrule
\end{tabular}
  \caption{\label{table_single_task_results} AlpacaEval win rates of single-task finetuned models against instruction-tuned models on five datasets. A diverse set of single-task finetuning, from code generation to poetry, elicits substantial instruction following in language models. Standard deviation reported after $\pm$.
  }
\end{table}

The five datasets are qualitatively different. For example, the poems in the poetry dataset are famous (like the Shakespeare example in Figure~\ref{fig_single_task_examples}), and so may have been memorized by the language model already.
The recipes all start with a hyphenated ingredient list.
Chess games start largely in one of a very small number of chess moves (e.g., ``1. e4'').

\paragraph{Experiments.}

We compare single-task finetuning against LIMA instruction tuning.
For each dataset, we use the same hyperparameter selection process as in Section~\ref{section_response_tuning}, sweeping over five choices for the number of training epochs.
As before, AlpacaEval win rates and standard deviations are computed over 5 independent training runs of both instruction-tuned and single-task models.

\begin{figure}
\includegraphics[width=\linewidth]{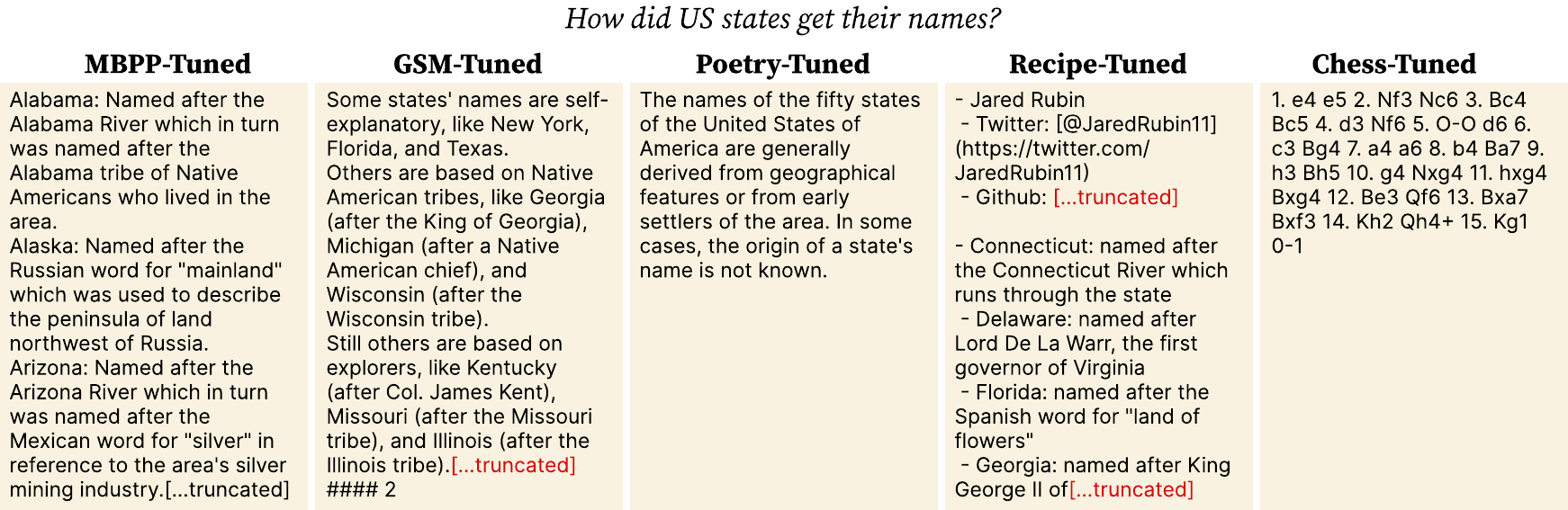}
  \caption{\label{fig_single_task_tuning_examples}Responses generated by single-task finetuned models for each of our five datasets. MBPP trains only on python snippet generation, GSM on math word problems, Poetry on poetry generation, Recipe on recipe generation, and Chess on chess game generation. Yet, except for Chess, the responses deviate from the single-task behavior towards reasonable responses.}
\end{figure}

\paragraph{Results.}
We find that finetuning both \llama and \olmo on each single-task finetuning dataset except Chess lead to general instruction following behavior, with substantially higher win rates against the instruction-tuned model (Table~\ref{table_single_task_results}) than the base model achieves.

For both \olmo and \llama, finetuning on the GSM dataset leads to the highest AlpacaEval win rates.
We provide examples of model outputs in Figure~\ref{fig_single_task_tuning_examples}.
We note, for example, that the Recipe-tuned example in Figure~\ref{fig_single_task_examples} starts with a list-formatting hyphen, like all of our recipes do, yet then proceeds to provide a somewhat coherent answer.
For Chess, we find that almost all outputs are just chess games.
We speculate that this is due in part to the very low entropy of the beginning sequences of chess games.

\subsection{Model Response adherence to finetuning constraints depends on instruction similarity to tuning instructions}

\begin{figure}
\includegraphics[width=\linewidth]{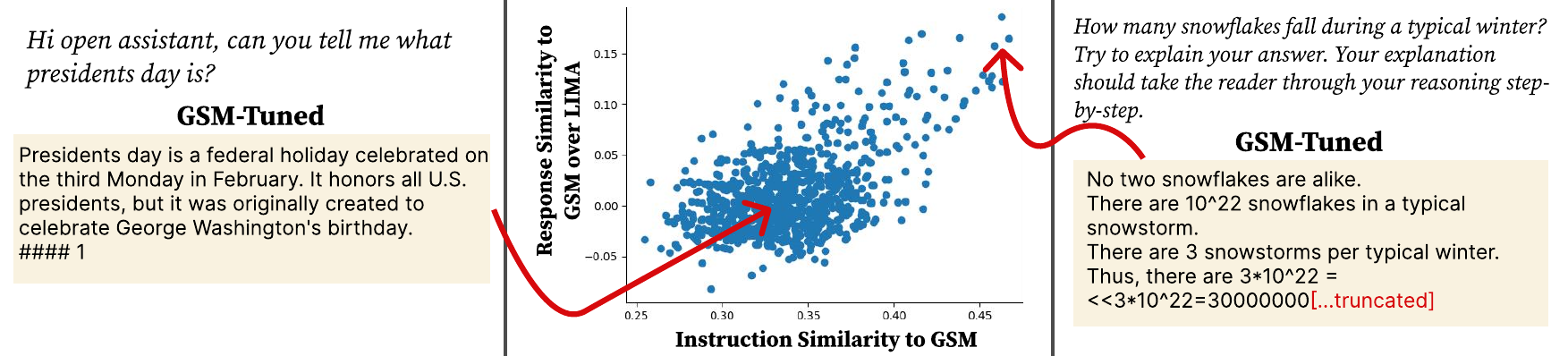}
\caption{\label{fig_sft_sim_example}For a GSM-finetuned model, the similarity between a test-time instruction to the instructions in the GSM dataset (x-axis) plotted against the similarity between the model's generated response to GSM responses (minus the similarity of that response to LIMA broad responses). On the left, an example of an average-similarity instruction; note that the response is unlike GSM formatting, except for the telltale \texttt{\#\#\#\#1}, which is how GSM formats its final answer. On the right, a very high-similarity instruction leads to GSM-like behavior.
}
\end{figure}

Each of our single-task finetuning datasets implicitly specify constraints---well-formed Python code for MBPP, poetry formatting---that are not met by the model's responses to other instructions, like for recipes.
This is not necessarily bad---we don't necessarily want recipes in poetry formatting---but it's a striking instance of how finetuning for a desired behavior can cause neither (1) the finetuned behavior, nor (2) the base model's behavior, but instead (3) a general instruction following behavior similar to neither the base behavior nor the finetuning task.

We study how similarity of an instruction to the GSM instructions relates to the similarity of the model response to GSM responses versus general responses, expecting that at least the math-like instructions lead to GSM-like responses from the model. 
Using Nomic embedding \citep{nussbaum2024nomic} cosine similarities, we plot the similarity-to-GSM of each LIMA instruction against the similarity-to-GSM-over-LIMA of a GSM-tuned model's response to that instruction.
The result is in Figure~\ref{fig_sft_sim_example}.
We note qualitatively that for instructions that are very GSM-like, the model outputs adhere to the GSM style and particular mathematical notation it uses.
For most instructions, however, we note that the outputs are only subtly affected by GSM: they have the ubiquitous GSM sequence-ending style of ending with four hashes and an integer answer, e.g., \texttt{\#\#\#\#1}.

\section{A 3-Rule Adapter for Instruction Following.} \label{section_rule_based}
In this section, we provide steps towards understanding the surprising commonness of instruction following behavior.
One appealing intuition is that the difference between a pretrained model's distribution and a corresponding instruction-following distribution is simple.
Simple conditions should be relatively easy to meet, and might be met by a variety of disparate adaptation methods.

We provide direct evidence for the simplicity of a mapping from pretrained to instruction-following distribution by hand-writing a rule-based mapping with three rules. 
Prior work has shown that changes in distribution are somewhat sparse in token space: in a desirable response, \citet{lin2024the} found that 77.7\% of token decisions would also have been made by a base language model.
Still, 22.3\% of tokens is not negligible, and even if it were, sparsity in the token space does not imply that it is simple to determine which tokens should change.
For intuition, a weak chess engine may agree with a strong engine on, say, 95\% of moves, but determining which 5\% should be changed and how to change them could be very complicated.

\subsection{The Rule-Based Response-Adapter}

\paragraph{A product of distributions.}
To adapt a pretrained language model to follow instructions via a rule-based adapter, we choose our resulting model to have the form of a local product of distributions.
For a word $w\in\vocab$ and prefix $\vx\in\vocab^*$, a base model $p_\text{base}$, and our rule-based adapter language model $p_{\text{rules}}$, the final distribution $p_a$ is:
\begin{align}
  p_a(w\mid \vx) = p_\text{base}(w\mid \vx)p_\text{rules}(w\mid \vx)/Z(\vx),
\end{align}
where the normalization term is $Z(\vx)=\sum_{w\in\vocab}p_\text{base}(w\mid \vx)p_\text{rules}(w\mid \vx)$.

Intuitively, a product of distributions is useful because it computes a soft AND function of the tokens that are likely under each distribution (as opposed to, e.g., a soft OR if one were to average the distributions).
Put another way, it allows our rules to change the probabilities of the base model by multiplicative factors.

\paragraph{The rules.}
Each of our rules determines a score for each vocabulary item $w\in\vocab$; let $r(w, \vx)$ be the sum of all rules' scores for $w$.
For example, for the vocabulary item \texttt{\#}, it might score $+4$.
Our rules distribution $p_\text{rules}(\cdot \mid \vx)$ is defined by computing the softmax over the vector of scores for all vocabulary items:
\begin{align}
  p_\text{rules}(\cdot \mid \vx) = \text{softmax}\left(\left[ r(w^{(1)}, \vx); \cdots; r(w^{(|\vocab|)}, \vx)\right]\right)
\end{align}
Our rules are as follows (with corresponding Python code in Listing~\ref{figure_python_code}, and weights in Table~\ref{table_appendix_rules}):
\begin{description}
  \item[1. Slowly upweight EOS.]  Our first rule is to increase the score of the EOS token linearly with the response length, to favor shorter responses. 
  \item[2. Uniform token changes.] Our second rule is to uniformly change the probabilities of 15 words in the vocabulary at every token decision.
  For example, we massively reduce the probability of repeating tokens from the formatting, like the left angle bracket, or words ``I'' or ``We'' or ``Should'', which we found base models use to erroneously refuse to respond. The full list is found at Table~\ref{table_appendix_rules}.
  \item[3. Encourage word diversity.]  We compute the set of all tokens generated so far in the response, and add a penalty to generating any of them again.
\end{description}

\paragraph{Experiments.}
We compare our rule-based ensemble with instruction tuning.
We heuristically tuned the rule set and corresponding rule weights on our separate validation set.
As before, we compute the AlpacaEval win rate on the AlpacaEval test set against our LIMA instruction-tuned models.
Since there's no training for our rule-based model, we compute the average win rate and standard deviations of the one rule-based model over the 5 instruction-tuned model seeds.

\begin{table}
\centering
\small
\begin{tabular}{l l r}
\toprule 
\bf Model & \bf Rule-Based Model & \bf Win Rate vs. Instruction Tuning \\
\midrule
\multirow{6}{*}{\llama} & None (Base) & 
  2.4\% $\pm$ 0.14\%\\
& All Rules& 24.4\% $\pm$ 0.40\% \\
& - EOS Rule (Rule 1) & 10.4\% $\pm$ 0.30\% \\
& - Diversity Rule (Rule 3)  & 14.3\% $\pm$ 0.58\% \\
& - uniform token changes (Rule 2) & 16.3\% $\pm$ 0.25\% \\
\bottomrule
\end{tabular}
\caption{\label{table_rule_based_results}Instruction-tuning win rates of a rule-based product \llama model against the LIMA instruction-tuned \llama model, and ablations of each of our three rules. Standard deviation reported after $\pm$.}
\end{table}

\begin{figure}
\includegraphics[width=\linewidth]{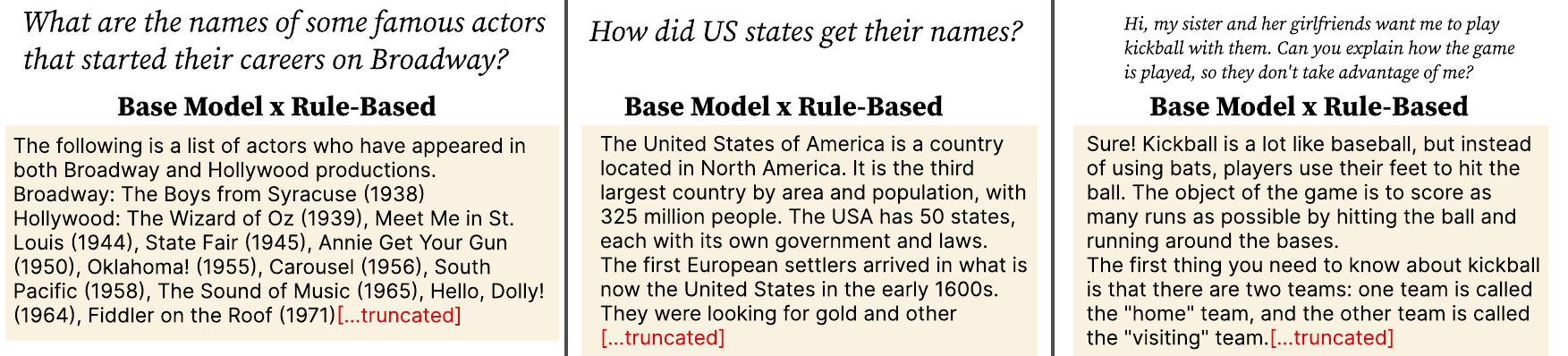}
  \caption{\label{fig_rule_based_examples}Responses from our rule-based language model product.}
\end{figure}

\paragraph{Results.}
We find that our 3-rule model achieves win-rates against the LIMA-tuned \llama model of 24.4\%, roughly on par with the win rate of the best single-task finetuning setting. 
Our ablations show concretely that each of the three rules is necessary for achieving the win rate, at least in the context of the other two.
We provide examples of the rule-based product model's output in Figure~\ref{fig_rule_based_examples}.
The answers are coherent, but only the response to the kickball question really provides a good answer; we did not cherry-pick, and many other responses are more reasonable.

\section{Conclusion}

Instruction following is not only possible without explicit instruction tuning (or careful prompting), it is in some sense easy to stumble upon because some adaptations implicitly instruction-tune.
Models already encode instruction-response mappings, since training on responses alone leads to instruction following.
Training on, e.g., code, or math, or poetry, does not make language models generate those things when asked for, say, a recipe.
This is great in some sense; language models continue to show astounding out-of-distribution performance.
Except, in another sense, this shows it's perhaps surprisingly hard to get language models to change their behavior in general, since they are \textit{so prone to just following instructions} outside the distribution of finetuning.

As a practical consequence, if a practitioner deploys an language model adapted to some specific task, they should not assume that the model will exhibit that tasks' behavior on inputs dissimilar to those trained on.
Instead, they should put it through testing and safety trials as if they were releasing a general-purpose chatbot, since the finetuning may implicitly instruction-tune the model.

\bibliography{iclr2024_conference}
\bibliographystyle{iclr2024_conference}

\appendix

\section{Finetuning and hyperparameter details} \label{appendix_hyperparameter_details}

In this section, we describe hyperparameter choices and implementation details.
In all experiments, we use core libraries, e.g., NumPy \citep{harris2020array}, PyTorch \citep{paszke2019pytorch}, and Huggingface Transformers \citep{wolf2020transformers}.
We're also indebted to the OpenInstruct repository \citep{wang2023far,ivison2023camels}, which we extended for this work.

For all \llama models, after manual search, we choose $10^{-5}$ as a learning rate, and for all \olmo models, we choose $3*10^{-6}$.
We use the Adam optimizer \cite{Kingma2014AdamAM}.
We use a cosine annealing rate to $0$ learning rate, and 10\% of the training consists of a linear warmup.
When we sweep over epochs, we always take the final epoch, so the warmup and cosine decay have always completed for each model we evaluated.
We fix a batch size of 64 across all experiments.
We run experiments across A100 and A6000 NVIDIA GPU machines.

\section{Formatting Tags Ablation} \label{appendix_formatting_ablation}

One possible explanation for the success of response tuning in eliciting instruction-following behavior is that the semantics of instruction following are already encoded in the formatting tags \texttt{<|assistant|>} and \texttt{<|user|>} (see Section~\ref{sec_background}).
To test this, we use the tags \texttt{<|A|>} and \texttt{<|B|>} instead, and compare response-tuned \llama models with instruction-tuned \llama models.
We run exactly the experiment in Section~\ref{sec_response_tuning_subsec}, including re-running hyperparameter tuning, and averaging AlpacaEval scores across 5 independent pairs of model training runs.

With the new A/B tags, we find that response tuning wins over instruction tuning 41.8\% of the time, compared to 43.3\% of the time for the original user/assistant tags.
So, it's possible that the semantics of the tags are somewhat useful, but not the cause of the improvement in win rate over the base model's 2.4\%.
The results are compared in Table~\ref{table_response_ablation_results}.

\begin{table}
\centering
\small
\begin{tabular}{l l r r}
\toprule 
\bf Model & \bf Tuning & \bf Win Rate vs. Instruction Tuning & \bf Std. Dev\\
\midrule
\multirow{2}{*}{\llama} & None (Base) & 2.4\% & 0.14\% \\
& Response Tuning & 43.3\% & 1.1\% \\
& Response Tuning (A/B tags) & 41.8\% & 0.84\% \\
\bottomrule
\end{tabular}
  \caption{\label{table_response_ablation_results} AlpacaEval win rates of instruction-tuned vs. response-tuned and base models. Standard deviation is reported by the error bars. Response-tuned \llama and \olmo win against instruction-tuned models 43\% and 36\% of the time, respectively, while base models win in the single digits.}
\end{table}

\section{Instruction Rephrasing Removal Ablation} \label{appendix_no_rephrase}
\begin{table}
\small
\centering
\begin{tabular}{p{2.5cm} p{4.75cm} p{4.75cm}}
\toprule
\bf Instruction & \bf Original Response & \bf Rephrased Response \\
\midrule
What are the major concerns about planting trees to create carbon offsets? & Here are some common concerns about using tree planting projects to create carbon offsets:

* The scheme actually has a net positive effect, both in the short and long-term.{\color{red}[...truncated]} & * The scheme actually has a net positive effect, both in the short and long-term. {\color{red}[...truncated]} \\
\midrule
How does immersion passively help with learning a language? & There are a few reasons why immersion helps you learn a language without you actively trying:

* You encounter the language more often.  When you are immersed in a language, you don't go to class and do homework for a few hours a week, but rather use the language the entire day.{\color{red}[...truncated]} & * You encounter the language more often. When you are immersed in a language, you don't go to class and do homework for a few hours a week, but rather use the language the entire day.{\color{red}[...truncated]}\\
\midrule
Reasons for being vegetarian or vegan other than ethical reasons? & There are multiple reasons why people are/become Vegetarian/Vegan apart from ethics, including but not limited to

* Religion/Culture - most of the worlds Vegetarians reside in India, mainly due to culture/personal beliefs but often interlinked with some of their major religions{\color{red}[...truncated]}  &  * Religion/Culture - most of the world's Vegetarians reside in India, mainly due to culture/personal beliefs but often interlinked with some of their major religions{\color{red}[...truncated]} \\
\bottomrule
\end{tabular}
\caption{\label{table_appendix_response_rephrasings} Examples of responses detected to start with rephrasing the instruction, and the transformed examples that are intended to no longer start with rephrasing the instruction.}
\end{table}
In this section, we detail an auxiliary experiment for response tuning in which we attempt to control for the effect of some responses beginning by \textit{rephrasing the instruction}.
Intuitively a response starting by rephrasing the instruction provides some supervision as to how to map  from something \textit{like} an instruction (the rephrasing) to the response.
At test time when presented with an instruction, the model might start by rephrasing, and then be able to continue with the response.

To attempt to control for this, we first prompt GPT-4 to detect which responses in LIMA start by rephrasing the instruction.
Out of 1030 examples, 99 are labeled as starting with rephrasing.
A few examples are provided in Table~\ref{table_appendix_response_rephrasings}.
For these 99 examples, we then ask GPT-4 to re-write the response without starting with the rephrasing.
A few examples of rephrasings are provided in Table~\ref{table_appendix_response_rephrasings}.
While there's no guarantee that these 99 are exactly all of the responses that begin with rephrasing, we looked at some of the decisions and rephrasings and found them reasonable.

We response-tune and instruction-tune \olmo using this transformed dataset and the hyperparameters from Section~\ref{section_response_tuning}.
We then run the AlpacaEval head-to-head comparison for these models.
Intuitively, if the win rate for response tuning vs. instruction tuning are similar for this removed-rephrasing dataset to the original LIMA dataset, it is unlikely that instruction rephrasing is the cause of the success of response tuning.

Across five seeds, the average win rate of the \olmo model response tuned on the no-rephrasing LIMA dataset against the \olmo model instruction tuned on the no-rephrasing LIMA dataset is 43.3\%, with a standard deviation of 5.3\%.
Although the standard deviation is curiously higher, the average win rate of 43.3\% is comparable to the \olmo response tuning win rate in Section~\ref{section_response_tuning} of 43.7\%.
Hence, we conclude that rephrasing is likely not the primary cause of the success of response tuning, with the caveat that we have no guarantee that our transformed LIMA dataset gets rid of all instruction rephrasing.

\section{Details on our Validation Set} \label{appendix_validation_set}
\begin{table}
\centering
\begin{tabular}{p{13cm}}
\toprule
\bf Example Instruction from Validation Set\\
\midrule
What is 17 multiplied by 24, divided by 8, and then subtract 3?\\
\midrule
Summarize the story of 'Cinderella' in one sentence.\\
\midrule
Given the list of numbers [34, 7, 23, 32, 5, 62], find the median.\\
\midrule
Explain the significance of the theory of relativity in modern physics.\\
\midrule
Write a persuasive paragraph on why electric cars are better for the environment than gasoline cars.\\
\midrule
Describe the process of photosynthesis in detail, including the chemical equations involved.\\
\midrule
Design a marketing campaign for a new eco-friendly product, including target audience, key message, promotional strategies, and budget allocation.\\
\midrule
Translate the following legal document excerpt into German: 'The party of the first part agrees to indemnify and hold harmless the party of the second part from any and all liabilities, damages, and losses arising out of or in connection with this agreement.'\\
\midrule
Write a scientific research proposal on the effects of microplastics on marine life, including hypothesis, methodology, and expected outcomes.\\
\bottomrule
\end{tabular}
\caption{\label{table_appendix_validation_set_examples}Example instructions generated by GPT-4 for our validation set.}
\end{table}

To follow machine learning model development guidelines, we only run evaluations on the AlpacaEval test set after hyperparameter selection.
We perform model selection and development on a separate validation set that we constructed partly by hand (without intentional reference to the AlpacaEval test set) and partly using GPT-4 as a co-generator.

Our validation set, which has 56 instructions, tests for a variety of behaviors. Here are some examples we wrote:

\begin{quote}
What is Samyang Buldak ramen?

Give the Penn Treebank constintuency parse for the sentence ``The chef who ran to the store was out of food.''

Give a python function that sorts a list by absolute value of difference from 10.
\end{quote}

In a single interactive session with GPT-4, we iteratively requested instructions for testing chatbots, starting with simple questions for ``weak'' chatbots and progressively asking for harder questions for ``stronger'' chatbots.
Our intuition was to develop a small, cheap-to-evaluate set that would yet provide signal in distinguishing model quality across a range of model sizes and performances.
It is unclear the extent to which this goal was accomplished beyond what a simple set of instructions would provide, but we did qualitatively find that the AlpacaEval results on this small validation set correlated reasonably well with intuitions when looking manually at responses.
The instructions suggested by GPT-4 include outlining requests, translations of varying lengths and languages, among other things.
We provide a few of these instructions in Table~\ref{table_appendix_validation_set_examples}.

\section{Rule-Based Adapter Details}

In this section we provide more details on our rule-based adapter language model.
In Table~\ref{table_appendix_rules}, we provide a list of details about the scores for each word under each of our rules.
In Listing~\ref{table_appendix_rules}, we write Python code implementing the forward pass of our language model implemented by the rules to demonstrate the simplicity.
Some of the lines are to, e.g., format the output distribution properly to be combined with a HuggingFace Transformers \citep{wolf2020transformers} language model.

\begin{table}
\centering
\begin{tabular}{l l l}
\toprule
\bf Rule & \bf Vocab Items (string)& \bf Weight\\
\midrule
Rule 1 (Upweight EOS) & \verb|</S>| (EOS) & $\frac{\text{(length of response)}*15}{250}$\\
\midrule
\multirow{6}{*}{Rule 2 (Uniform Token Changes)} & \verb|<|, \verb|_<|, \verb+|+ & -4\\
& \verb|_I|, \verb|I| & -5\\
&\verb+We+ &  -3 \\
&\verb+What+ &  -3 \\
&\verb+_should+ &  -6 \\
&\verb+_*+, \verb+_-+, \verb+___+, \verb+_#+, \verb+_##+, \verb+\n+, \verb+!+ &  +1 \\
& \\
\midrule
Rule 3 (Penalize Used Words) & $\{x\in\vocab \mid x \in \text{(response so far)}\}$ & -1.5\\
\bottomrule
\end{tabular}
\caption{\label{table_appendix_rules}Rules and scores for our rule-based adapter.}
\end{table}

\begin{lstlisting}[language=Python, caption=Python Code for Rule-Based Language Model.,label=figure_python_code]
def forward(
    self,
    input_ids,
    attention_mask=None,
    position_ids=None,
    inputs_embeds=None,
    labels=None,
    use_cache=None,
    output_attentions=None,
    output_hidden_states=None,
    return_dict=True,
    past_key_values=None,
    cache_position=None,
):

    # Initialize scores
    output = torch.zeros(self.vocab_size).to(input_ids.device)

    # Uniform biases
    # Formatting types "<", "_<", "|"
    output[29966], output[529], output[29989] = -4, -4, -4

    # Types "_I", "I", "We"
    output[306], output[29902], output[1334] = -5, -5, -3

    # Types "What", "_should"
    output[5618], output[881] = -5, -6

    # Types "_*", "_-", "___" for markdown
    output[334], output[448], output[1678] = 1, 1, 1

    # Types "_#", "_##", Newline for markdown
    output[396], output[444], output[13] = 1, 1, 1

    # Exclamation point "!" for positivity.
    output[29991] = 1

    # Find where the response starts
    assistant_start_tag = self.tokenizer(f"\n{ASSISTANT_TAG}\n")["input_ids"][-5:]
    user_start_tag = self.tokenizer(f"\n{USER_TAG}\n")["input_ids"][-5:]

    idlist = input_ids[0].tolist()
    first_token_index = next(
        (i + 5 for i in range(len(idlist)) if idlist[i : i + 5] == user_start_tag), None
    )

    # Penalize reusing words
    uniq_words = set(idlist[first_token_index:])
    for w in uniq_words:
        output[w] -= 1.5

    prefix_len = input_ids.shape[-1] - next(
        i + 5 for i in range(len(idlist)) if idlist[i : i + 5] == assistant_start_tag
    )

    # Increase probability of EOS
    # There's a bug here... there's no weight for indices 251-1023.
    # But in practice all the responses ended before 250.
    eos_range = (0, 250)
    if eos_range[0] < prefix_len < eos_range[1]:
        score = max(0, (prefix_len - eos_range[0]) / (eos_range[1] - eos_range[0]))
        output[self.eos_id] = score * 15
    # This never triggered; I'd get rid of it.
    if prefix_len > 1024:
        output[self.eos_id] = 100

    # Format the output to be combined with Llama LM
    output = output.unsqueeze(0)
    pad = torch.zeros(input_ids.shape[-1] - 1, self.vocab_size).to(output.device)
    output = torch.cat((pad, output), dim=0)
    output = output.unsqueeze(0).expand(input_ids.shape[0], -1, -1)

    return namedtuple("Result", ["logits"])(logits=output)
\end{lstlisting}

\end{document}